\documentclass[journal]{IEEEtran}
\usepackage{graphicx}
\usepackage{amsmath,amssymb} 
\usepackage{color}
\usepackage{subfigure}
\usepackage{subfloat}
\usepackage{booktabs}
\usepackage{algorithm}
\usepackage[noend]{algpseudocode}
\usepackage{hyperref}

\newcommand{\etal}{et al. }

\begin{document}
\title{Geometric Scene Refocusing}

\author{Parikshit Sakurikar and P. J. Narayanan\\
Center for Visual Information Technology - Kohli Center on Intelligent Systems,\\
International Institute of Information Technology - Hyderabad, India\\
{\tt\small \{parikshit.sakurikar@research.,pjn@\}iiit.ac.in}
}

\maketitle

\begin{abstract}
An image captured with a wide-aperture camera exhibits a finite depth-of-field, with
 focused and defocused pixels. A compact and robust representation of focus and defocus
 helps analyze and manipulate such images. In this work, we study the fine characteristics
 of images with a shallow depth-of-field in the context of focal stacks. We present a 
composite measure for focus that is a combination of existing measures. We identify 
in-focus pixels, dual-focus pixels, pixels that exhibit bokeh and spatially-varying 
blur kernels between focal slices. We use these to build a novel representation that 
facilitates easy manipulation of focal stacks. We present a comprehensive algorithm 
for post-capture refocusing in a geometrically correct manner. Our approach can refocus 
the scene at high fidelity while preserving fine aspects of focus and defocus blur.
\end{abstract}


\IEEEpeerreviewmaketitle

\section{Introduction}
Sharp and soft focus are important attributes of a good photograph.
Focus and defocus blur are used creatively by photographers to produce
remarkable compositional effects. An image captured using a wide-aperture camera
is a collection of differently focused scene points.
Cameras with a wide aperture are available on devices ranging from high-end mobile
phones to DLSRs. The relative geometry of the sensor and the lens 
at capture-time governs the points in the scene that appear focused in the image.
Light arriving from these focused points contributes to a single or very few pixels on the sensor. 
Other regions of the scene appear defocused by an amount proportionate to their distance 
from the in-focus region. The luminosity of defocused scene points is distributed across 
a set of proximate pixels, and this spread is referred to as a defocus kernel.
The size and shape of this kernel depends on the distance
of the scene point from the in-focus region and its 2-dimensional pixel location
in the image. Defocus kernels of proximate pixels may overlap with each other, leading
to complex pixel interactions. 
An accurate model of focus and defocus blur
is relevant to computational photography as it can 
enable measurement of focus for tasks such as de-blurring, 
depth-of-field extension, refocusing and depth-from-focus. 
Post-capture modeling of focus and defocus blur using only a single image is
an ill-constrained problem. Accurate focus modeling usually requires multi-focus imagery
and a robust method to estimate in-focus pixels and defocus kernels.

In-focus pixels can be estimated by measuring the sharpness across each pixel's neighborhood.
In-focus pixels are expected to be sharp while defocused pixels exhibit low contrast.  
However, sharpness alone is an unreliable estimate as it is dependent 
on the texture and arrangement of scene points. Furthermore, the intensity of a pixel is 
quantized by the dynamic range of the sensor and may not represent
the original luminosity of its scene point. The intensity might also be mixed
with the defocused intensity of other scene points. These issues complicate the task
of identifying whether a pixel is in focus and by what amount. 

Modeling the defocus kernel at a pixel is also a challenging task.
The size and shape of a defocus kernel depends on the camera 
and scene geometry. The kernel shape is also affected by vignetting
or kernel-shortening close to the boundaries of the aperture.
The defocus spread from a farther scene point may be partially
occluded by the presence of closer objects. Such interaction between defocus kernels
from different depths requires accurate geometric modeling.
In this work, we propose a robust method to estimate and manipulate focus and defocus blur.
The main contributions of this work are:

\begin{enumerate}
\item We identify a composite measure of focus that combines the strengths
of existing focus measures for estimating the true intensity and in-focus
location(s) for each pixel.
\item We build a compact representation for multi-focus input imagery
using the composite measure and a calibration method that  
estimates the size and shape of defocus kernels.
\item We propose a novel algorithm for geometric depth-of-field manipulation.
Our algorithm correctly accounts for complex pixel interactions
at depth edges using an occlusion coefficient.
\end{enumerate}
Ours is a comprehensive study of focus and defocus and is applicable to
variety of scenes captured as a focal stack.
Careful analysis of focus and defocus blur along with simple models 
that help in synthesis are the primary strengths of this work. 
We expect our model to be used by image editing tools in conjunction with 
focal stacks or RGBD images to allow easy and quick post-capture 
manipulation of focus.

\section{Related Work}
Focus and defocus blur have been widely studied in computer vision in the past, primarily in the
context of image de-blurring to create an in-focus image of the scene.
Since it is difficult to estimate the per-pixel focus profile from a single image,
two or more images of the scene with different focus positions are typically used to
measure focus and defocus blur. Focus variations in an image also provide implicit cues to 
estimate scene depth.
We broadly discuss the contemporary work in computational photography that deals with 
wide-aperture images.

\paragraph{\textbf{Epsilon Focus Photography - Focal Stacks}}
A focal stack is a collection of multiple wide-aperture images with a 
small change in the focus position between consequent shots.
Focal stacks enable dense modeling of the focus profile for each pixel. 
These focus profiles are useful for tasks such as de-blurring, refocusing and depth estimation.
Hasinoff \etal \cite{Hasinoff,HasinoffLightEfficient,Kutulakos} show that focal stacks
require a significantly less time to capture and exhibit reduced noise characteristics 
compared to a single image capturing the full depth-of-field. This however leads to a
loss of temporal resolution due to the finite amount of time taken to capture each
slice. Focal stacks are thereby limited to mostly static scenes, as moving
objects complicate pixel-alignment across focal slices. Moreover, storing a full focal stack on 
the device can lead to large storage overheads on portable devices. Our previous work \cite{fsrpr} 
illustrates a compact representation for a focal stack that enables basic post-capture focus control.
In this work, we propose a representation that can reduce the storage complexity while also encoding 
the fine characteristics of each focal slice.

\paragraph{\textbf{In-Focus Imaging}}
Estimating the in-focus scene content from focal slices 
is the primary application of focal stacks.
An in-focus image of the scene is free of any ambiguity
caused by defocus blur for tasks such as segmentation, recognition and retrieval.
Kubota \etal \cite{KubotaMain} generate in-focus images using
linear filtering of focal slices using a focal-texture model.
Kodama and Kubota \cite{Kubota2013} extend this to a 3D
filtering method for generating dense pin-hole views from novel
viewpoints. Iterative computation of focal textures has also be used in our previous work
to produce in-focus images for shallow focal stacks \cite{Sakurikar}. Nagahara \etal
\cite{Nagahara} use a special camera to capture a focal sweep of the scene by
aggregating the light on the sensor during relative translation between the sensor
and the lens.  Deconvolution of the blurred focal sweep image
yields an all-in-focus image.  Kuthirummal \etal \cite{Nayar11}
use the same principle and average the captured focal slices to generate the
integrated image from a focal stack.  Agarwala \etal \cite{Agarwala}
generate an in-focus image from a focal stack using a global
contrast maximization approach. In this work, we comprehensively identify
the in-focus scene content for all pixels, including pixels that might exhibit multiple
in-focus locations. This occurs at pixels from background segments that might be
partially occluded by the foreground. We also locate the intensity saturated regions
in the scene and approximately estimate the in-focus intensity for these scene points.

\paragraph{\textbf{Measuring Focus}}
A focus measure is at the heart of any approach that uses wide-aperture images. 
Estimating whether a pixel is in focus or not requires a suitable focus measure.
Measures of focus based on several image properties have been proposed in computer vision
literature \cite{Nayar,Helmli,Subbarao,ShenChen,YangNelson}.
The response of a focus measure not only depends on the parameters of the measure
but also on the textural content around the pixel in the image.
Measuring per-pixel focus based solely on a single pixel's response
is usually noisy and unreliable \cite{ZhouTR12}.
Smooth focus maps that consider neighborhood consistency
can be generated using optimization methods such as Cost-Volume Filtering \cite{CostVolume}
or MRF labeling \cite{Boykov}. 
Iterative optimization methods can also be used to estimate smooth focus maps
 \cite{Sakurikar,egsrAkira}. 
Pertuz \etal \cite{Pertuz} analyze and compare several focus measures independently 
for the task of depth-from-focus. They conclude that Laplacian based operators are best
suited under normal imaging conditions. In \cite{Variational},
the Laplacian focus measure is used to compare classical DfF energy 
minimization with a variational model. A Ring Difference Filter measure is proposed in 
\cite{RDF}, with a filter shape designed to encode the sharpness around a 
pixel using both local and non-local terms. Mahmood \etal \cite{Mahmood}
combine three well known focus measures (Tenengrad, Variance and
Laplacian Energy) in a genetic programming framework. Boshtayeva
\etal \cite{Boshtayeva} describe the benefit of using multiple focus measures 
together in an anisotropic smoothing framework to compute scene depth. 
Measuring the focus information of a scene is in fact analogous to computing relative
scene depth from multiple focused images \cite{Nayar,SR,Favaro,Weickert,Subhashis2001,Seitz}.
In this work, we learn a composite measure of focus as a weighted combination of 
several informative measures, building on our previous work on depth-from-focus \cite{cfm}.

\paragraph{\textbf{PSF Modeling and Estimation}}
Modeling the spatially varying defocus kernel in wide-aperture images is relevant
in the context of image de-blurring and scene refocusing.
Kee \etal \cite{Kee} remove spatially varying optical blur by estimating dense non-parametric
blur kernels across the image. They describe calibration and kernel fitting methods for
blur estimation. Shih \etal \cite{Shih} show a calibration technique to predict the 
lens point-spread-function (PSF) at arbitrary
depths using a calibrated PSF at a known depth. Tang and Kutulakos \cite{Tang} describe an 
analytical approach for blind image de-blurring and PSF calibration at other pixels from known PSFs 
at some pixels. Hach \etal \cite{Hach} show a dense calibration method for a high 
quality lens with depth-aware rendering to produce synthetic bokeh.
In this work, we present a simple calibration method that estimates the size 
and shape of defocus kernels using the estimated in-focus pixels. We also identify
the blur kernels at intensity saturated scene regions and render them appropriately while
refocusing.

\paragraph{\textbf{Flexible Depth-of-Field Imaging}}
Changing the focus position or the depth-of-field of an image after
it has been captured is a useful tool in photography.
Nagahara \etal \cite{Nagahara} demonstrate flexible DoF effects by capturing disjoint
focal sweeps through the scene and deconvolving the
aggregate image by the integration kernel. Half-sweep imaging
\cite{Matsui} has also been used on the same lines. Jacobs \etal
\cite{Levoy} describe a precise composition model to
generate composite images from the available focal slices using a 
defocus map that specifies the amount of target defocus at each pixel.
They use geometric integration of rays for free-form depth-of-field control. 
The Lytro lightfield camera \cite{Ng,Lytro} enables post-capture refocusing 
and interpolation by capturing a 4D lightfield at a limited resolution.
In this work, we present a geometric approach for post-capture focus manipulation.

In a related parallel effort, we have used data-driven methods for post-capture focus 
control. In \cite{refgan}, we propose an adversarial learning framework trained on focal stacks created
from light-fields to refocus the scene after it has been captured.
The task of refocusing is decomposed into disjoint operations of de-blurring and re-blurring. 
We also propose an adversarial learning framework for defocus magnification
\cite{defmag}. Data driven methods for focus control have only recently gained popularity.
Wang \etal \cite{deeplens} achieve post-capture control of depth-of-field by estimating the depth
map using a convolutional neural network trained over a large RGBD dataset. 
Re-blurring the image to a target depth-of-field is also learned in a data-driven manner, 
with separate stream for image features and blur kernels. 
Data-driven methods suffer from a serious lack of generalizability
in our experience. These methods may be able to capture the fine characteristics of wide
aperture images, but will require large amounts of specific data to be able to do so effectively.
In this work, we propose a geometric refocusing algorithm that can perform flexible DoF 
manipulation on any scene using multi-focus input imagery.

\section{Wide-Aperture Imaging}
\begin{figure}
\centering
\includegraphics[width=.35\textwidth]{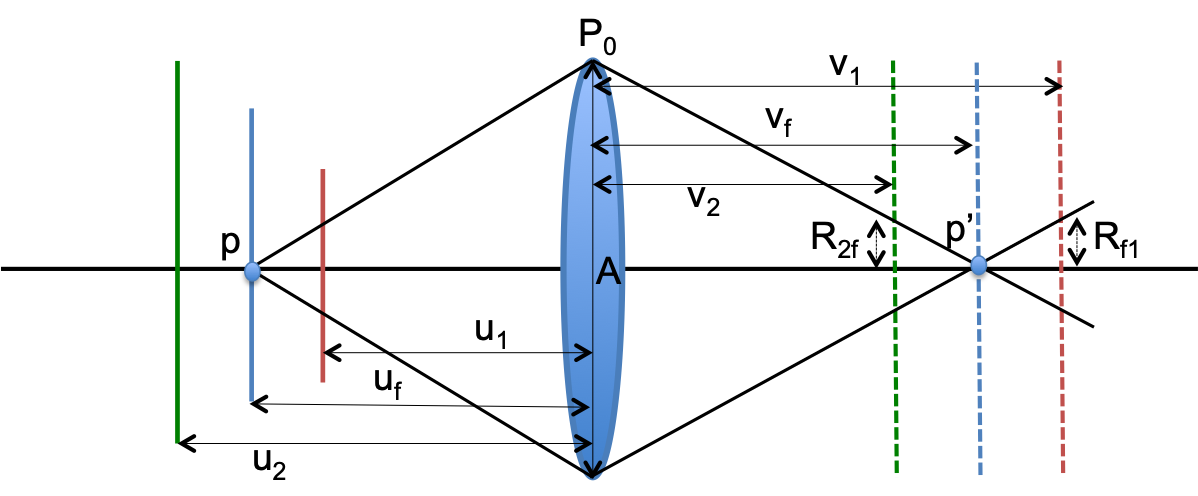}
\caption{
The rays from scene point $p$ at a depth $u_f$ converge at an in-focus pixel $p'$ when the sensor is positioned
at $v_f$. At other sensor positions such as $v_1$ and $v_2$, the rays from $p$ spread across a defocus
kernel of radius $R_{f1}$ and $R_{2f}$ respectively.}
\label{GeomRadii}
\end{figure}

An image captured through a finite aperture opening consists of a combination of focused
and defocused scene points. Unlike a pinhole camera that captures one 
(or very few) rays at each pixel, a wide aperture camera records the combination of several rays
at a pixel. If the rays at a pixel arrive from the same scene point, the point manifests as
an in-focus pixel in the image, while it appears as a defocused pixel if multiple scene points
contribute to it as illustrated in Figure \ref{GeomRadii}.

The characteristics of a wide-aperture image such as field-of-view, depth-of-field, 
focus distance, amount of defocus at each pixel, image brightness and sensitivity depend 
on the nature of the lens and the capture-time camera parameters. 
Image formation in wide-aperture cameras is typically modeled as a two stage process.
The first stage deals with the optical traversal of light through the lens assembly 
and its collection onto the discrete photo-electric sensor. The
second stage consists of the conversion of sensor voltages to discrete pixel
intensities in the image \cite{BrownICIP13Tutorial}.

Modeling the image formation pipeline in wide-aperture cameras
is important for accurate estimation and processing of focus and defocus blur.
In this work, we identify crucial
capture parameters such as in-focus pixels, dual focus pixels, defocus radii
and intensity saturation from multiple wide-aperture images captured as
an ordered focal stack.

A focal stack $\cal G$ is a sequence of $k$ images (called focal slices)
${\cal G}_i, 1 \le i \le k$.  Each slice is captured with a progressively varying
focal distance but the same aperture opening.
A focal slice ${\cal G}_i$ is the wide-aperture image corresponding to the focus position $i$
and can be defined as:
\begin{equation}
{\cal G}_{i} = \int\int h^i(x,y,d_{(x,y)}) \hat{{\cal G}}_r(x,y) \, dx\,dy \, ,
\label{fs}
\end{equation}
where $h^i$ is a spatially varying defocus kernel whose size and shape depends on the spatial location
of the pixel and the depth $d_{(x,y)}$ of its corresponding scene point and $\hat{{\cal G}}_r$ is 
the radiance or luminosity of the scene point.
An ideal focal stack captures each scene point in sharp focus in one and only one focal slice.
Focal slices exhibit magnification differences as the focus position varies.
This can be corrected using image registration or image alignment. Once registered,
a focal stack is a volume of pixels, with each node in the volume representing the pixel's 
intensity in the corresponding slice. The variation of pixel intensities across this
volume can be used to estimate in-focus scene points and defocus kernels as demonstrated
in the forthcoming sections.

\section{Measuring Focus}
\begin{figure}
\centering
\includegraphics[width=.45\textwidth]{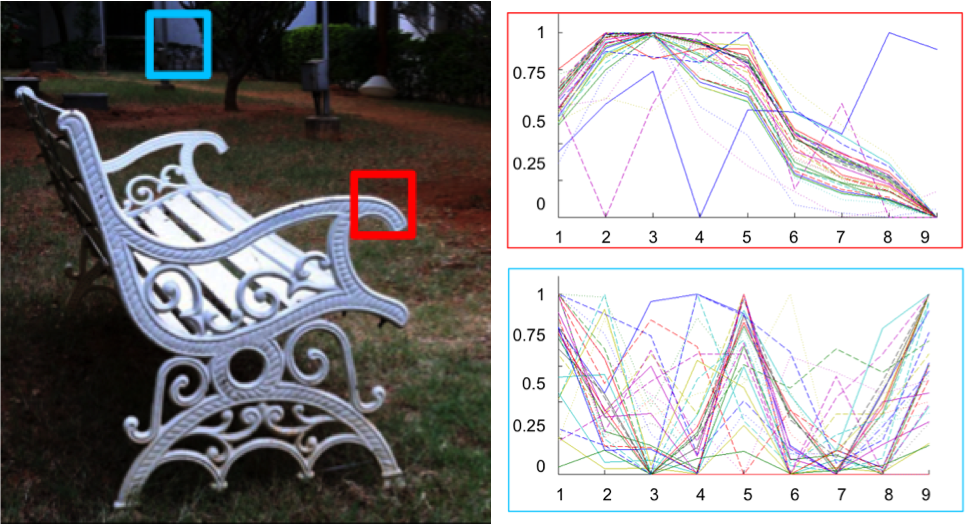}
\caption{The focus profiles of thirty different focus measures evaluated at two pixels 
across a focal stack. Severe variability in in-focus estimation across different
measures can observed based on the textural content around challenging pixels.}
\label{focusProfile}
\end{figure}

The sharpness of the image content across the two-dimensional neighborhood of a pixel is a reliable
indicator of its amount of focus. Identifying the appropriate measure of sharpness in a scene
independent manner is non-trivial. Figure \ref{focusProfile} shows the response of different measures
of focus/sharpness at two pixels in a focal stack. It can be seen that although these measures encode the
sharpness around a pixel, they exhibit remarkable variability.
A key contribution of our work is a composite focus measure (cFM) as a
weighted combination of existing measures.
 
We study the performance of 39 focus measures - all measures from Pertuz \etal \cite{Pertuz}, 
two additional measures from Boshtayeva \etal \cite{Boshtayeva} and the Ring Difference Filter from
\cite{RDF} - in the context of per-pixel focus estimation over a large dataset of focal stacks. 
We represent the focus measures using a similar convention to 
\cite{Pertuz}; the additional measures are labeled as HFN (Frobenius
Norm of the Hessian), DST (Determinant of Structure Tensor) and RDF (Ring Difference Filter).
We compute focus profiles shown in Figure \ref{focusProfile} for all the pixels in
our focal stack dataset across all focus measures.
We seek to select the best subset of focus measures using these focus profiles.
A supervised approach to focus measure selection
is not possible as ground-truth focus maps of the scene are not available. 
Unsupervised feature selection is thereby the natural choice for identifying a 
composite focus measure.

At each pixel, the agreement of different focus measures about 
the peak location of focus is of importance. Agreement among
multiple measures suggests a consensus in identification of sharp content around the pixel.
However, measures with highly identical responses at all slices have 
redundant information and may not be very useful together, even though they agree about 
the focus peak. Thus, we seek to identify
focus measures that demonstrate {\em high consensus} but {\em low correlation}.

\subsection{Consensus of Focus Measures}
To measure the consensus among focus measures we first estimate the mean 
peak focus position at each pixel. The measures that peak within a small
neighborhood of this mean position are then considered to be in consensus with each
other. For the mean focus position, we use all measures to build a coarse
focus map using MRF based energy minimization \cite{Boykov}. 
The data cost $D_L(p)$ of labeling the mean focus position of a pixel $p$ to focal slice
index $L$ is computed as the sum of normalized measure responses at the pixel:
\begin{equation}
D_L(p) = exp\left(-{\sum_{j=1}^{N_F} \frac{{F}_j(p,L)}{\sum_{l} {F_j(p,l)}}}\right),
\label{mrfData}
\end{equation}
where $F_j(p,L)$ is the  $j^{th}$ focus measure applied at pixel $p$ for the
$L^{th}$ focal slice and $N_F$ is the total number of measures. 
A multi-label Potts \cite{Boykov} term is used to assign smoothness costs.

\begin{figure*}
\centering
\includegraphics[width=0.9\linewidth]{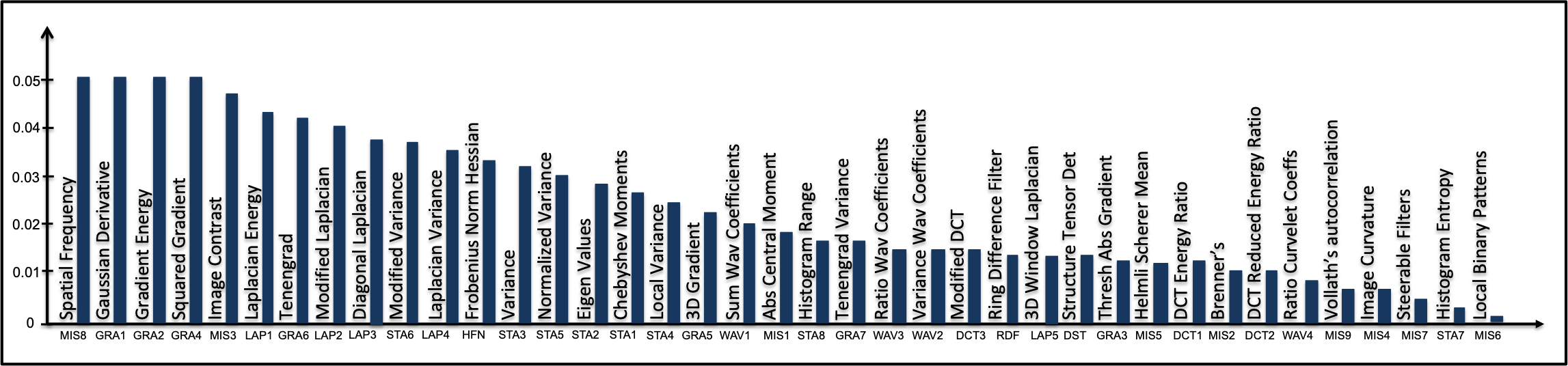}
\caption{A ranked list of all focus measures used in our analysis. The list is
sorted in descending order of the normalized cumulative consensus score $\hat{C}$
computed over our focal stack dataset.}
\label{ranking}
\end{figure*}

The result of MRF optimization for every focal stack is a neighborhood-aware mean focus 
position for all its pixels. 
The consensus for each focus measure is now recorded as the agreement of the measure
with this mean focus position within a small margin. The consensus $C$ is computed as
\begin{equation}
C(F_j;p) = \left\{ \begin{array}{cl}
1 & \textrm{if } \arg\max\limits_l F_j(p,l) \\
  & \in [i(p)-w,i(p)+w]\\
0 & \textrm{otherwise}
\end{array}\right\},
\label{cons2}
\end{equation}
where $i(p)$ is the label assigned after MRF optimization at pixel $p$ and 
$w$ denotes a small neighborhood around $i(p)$. We select $w$ as 10\% of
the number of focal slices in the stack. This corresponds to a small depth neighborhood
as the focus steps in our focal stacks are mostly uniform. $w$ may also be parameterized based on the blur
difference between two slices in case of non-uniform focus steps. The cumulative consensus score
for a measure is the sum of its consensus at pixels across all focal stacks used in our analysis:
\begin{equation}
\hat{C}(F_j) = \sum_{FS}\sum_pC(F_j;p),
\end{equation}
where FS represents all the focal stacks in the dataset and p represents the pixel in each stack.
Figure \ref{ranking} lists all measures used in our analysis, ranked according
to their normalized cumulative consensus score $\hat{C}$.

\begin{figure}
\centering
\includegraphics[width=0.8\linewidth]{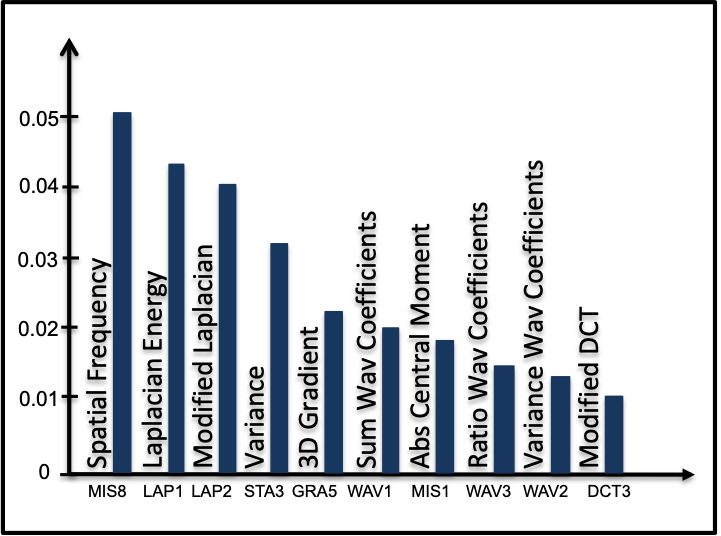}
\caption{Top 10 focus measures with a high degree of consensus but not high correlation.
The normalized consensus score is shown on the Y-axis. This score
is used as the weight for creating the composite focus measure.}
\label{consRank}
\end{figure}

\subsection{Correlation of Focus Measures}
The focus measures used in our analysis contain near-identical or highly correlated 
measures. These will naturally be in consensus with each other but add little additional 
value. Only one of each highly correlated
set of measures is a useful choice for the composite focus measure.
We compute pairwise correlation 
between focus measures across pixels of all focal stacks.
This is a cumulative correlation score and is computed as:
\begin{equation}
\hat{C}_r(F_i,F_j) = \sum_{FS}\sum_p\sum_l\sqrt{(F_i(p,l)-F_j(p,l))^2},
\label{corr}
\end{equation}
where FS indicates all focal stacks, p indicates all the pixels in a
focal slice and $l$ indicates the slices in the stack.
 
\begin{figure}
\centering
\includegraphics[width=0.55\linewidth]{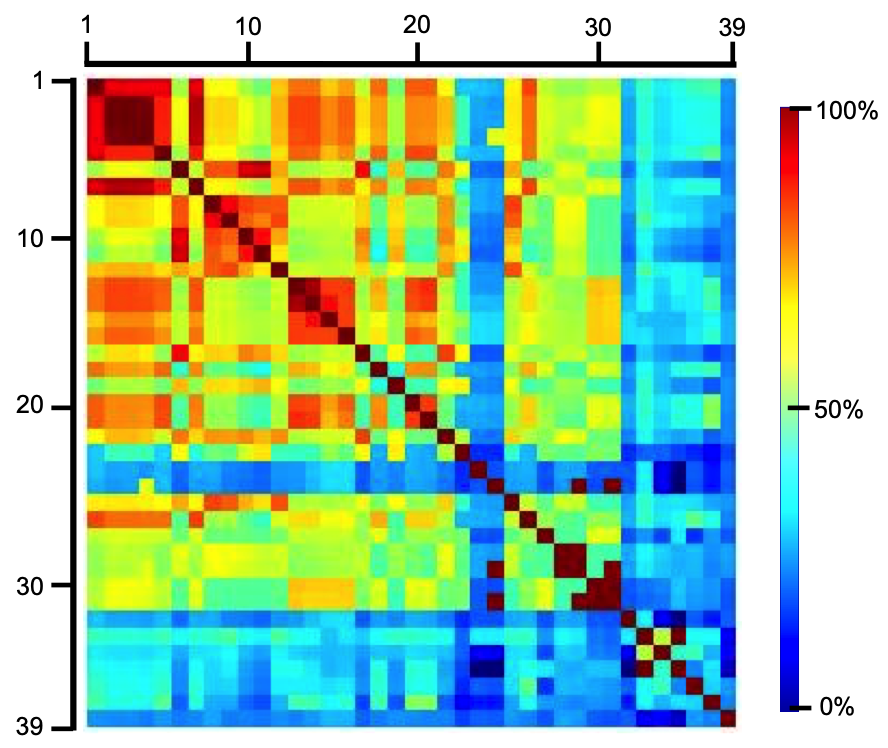}
\caption{Percentage pairwise correlation between pairs of focus measures (best viewed in color).
The FMs from left-to-right and top-to-bottom follow the same ordering shown in the 
ranked consensus list of Fig. \ref{ranking}.
FMs in dark shades of red indicate high correlation.
It is evident that the first five measures in descending order of consensus
are all highly correlated with each other. We would like to retain only one measure 
from these.}
\label{correlation}
\end{figure}

Figure \ref{correlation} shows the pairwise correlation between all
pairs of measures across the focal stack dataset.
From left-to-right and top-to-bottom, the measures 
are arranged in descending order of consensus $\hat{C}$. 
We seek to cluster these measures and only select one representative from 
each cluster. The similarity between focus
measures encoded in the correlation matrix of Figure \ref{correlation} 
can be used to cluster them.
We apply hierarchical agglomerative clustering on the distance matrix 
(reciprocal of the similarity matrix) to cluster correlated focus measures. 

We invert the correlation score between all pairs of measures to compute a distance
$d_{F_i,F_j}=\hat{C}_r(F_i,F_j)^{-1}$.
Agglomerative clustering is then applied with the single-linkage criterion,
which specifies the distance $D$ between clusters $c_i$ and $c_j$ as the minimum distance
between all the measures belonging to the clusters:
\begin{equation}
D(c_i,c_j)=\underset{F_i\in c_i,F_j\in c_j}{min}\,d_{F_i,F_j}
\end{equation}
The optimal number of clusters is computed using the gap statistic
\cite{gap2001} that uses the cumulative within-cluster distance $W_m$
for all candidate number of clusters m as
\begin{equation}
W_m= \sum^m_{n=1}\sum_{F_i,F_j\in c_n}d_{F_i,F_j},
\end{equation}
which is the sum of distances between measures belonging
to the same cluster. The optimal cluster count is set to the point
at which the logarithmic plot of the within-cluster distance $W_m$ exhibits an elbow or 
begins to flatten out. 

\begin{figure}
\centering
\includegraphics[width=0.85\linewidth]{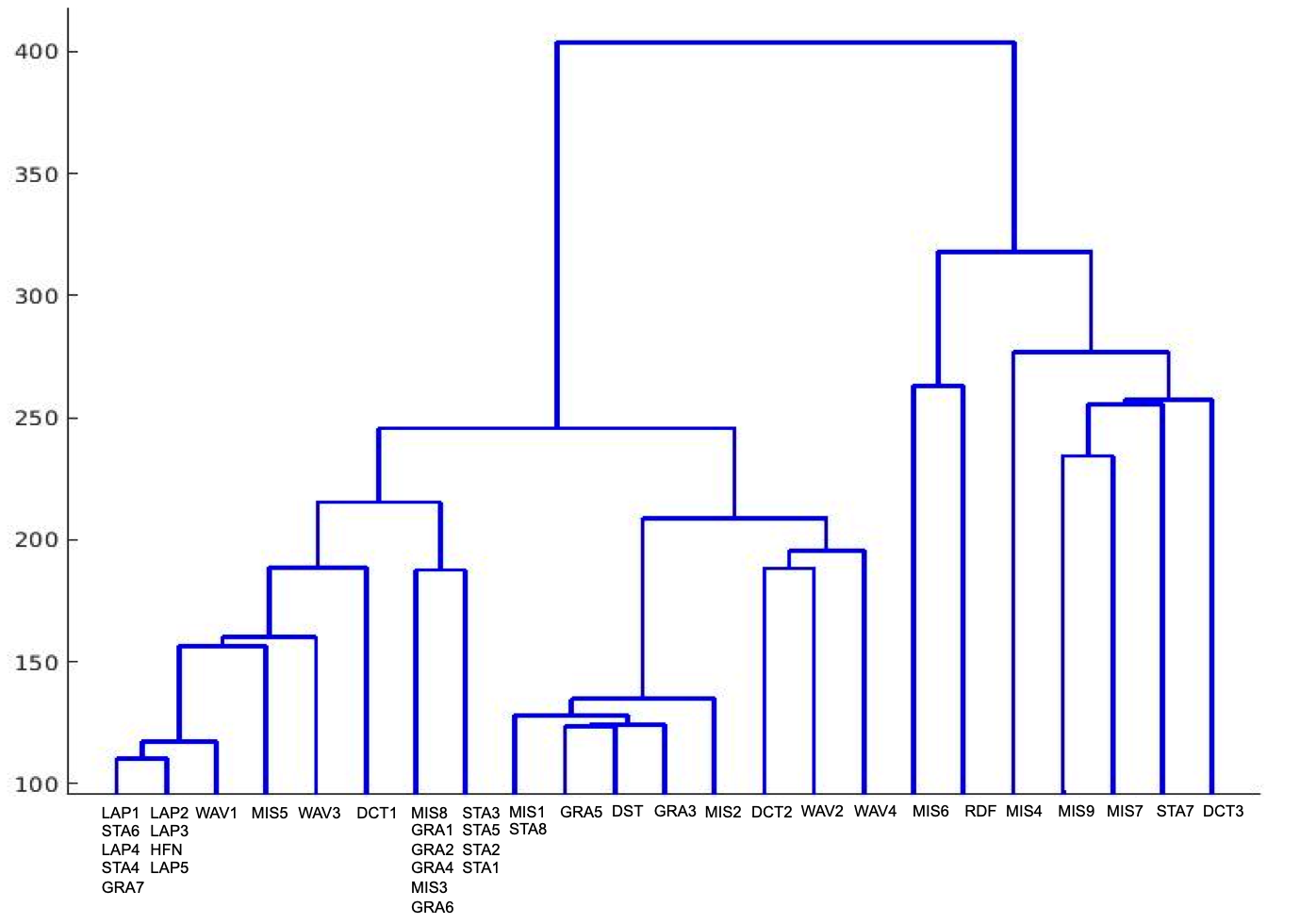}
\caption{23 Optimal clusters selected using the GAP statistic \cite{gap2001} applied 
with hierarchical agglomerative clustering on the correlation matrix of Figure \ref{correlation}.}
\label{dendro}
\end{figure}

We arrive at an optimal number of $23$ clusters for the $N_F=39$ measures used in our analysis. 
The dendrogram for this clustering is shown in Figure \ref{dendro}. 
We use the measure with the highest $\hat{C}$ as the representative measure for each cluster. 
We further sort all the representative measures based on $\hat{C}$. 
The top ten focus measures that exhibit high consensus but not high correlation
are shown in the list in Figure. \ref{consRank}. 

A weighted combination of the top five focus 
measures of Figure \ref{consRank} forms our composite focus measure (cFM).
The weight for each measure is its normalized cumulative consensus score. 
More measures from this list can be used for focus measurement at the cost of more computation
but the benefit of adding measures keeps decreasing as the list is traversed, as elaborated 
in \cite{cfm}.

\subsection{Evaluation of the cFM}
To verify whether the computed cFM is universal, we re-compute the consensus for all focus measures 
using Equation \ref{mrfData}, but the data term for each pixel is computed using
only the ten measures from the composite measure of Figure \ref{consRank}.
The correlation scores between the measures remain unchanged. We observe that there is 
no re-ordering in the composite measure on re-computing consensus scores, 
confirming the robustness of the computed composite measure. 

To confirm the generalization of the cFM, we evaluate the
consensus and correlation of focus measures for different subsets of our focal stack dataset.
We isolate four subsets containing 50 focal stacks each. The subsets are selected based on scene
attributes such as texture complexity, amount of blur, type of illumination
and random selection. 
Table \ref{nochangeTable} shows the top-five composite measure for the four subsets
of (1) scenes with high texture, (2) scenes with
high degree of blur, (3) outdoor scenes with bright illumination, 
(4) random subset of 50 focal stacks.
Such a categorization has little impact on the composite measure, the only
difference being that the HFN measure does not cluster within the LAP2 family for 
densely textured scenes. Overall, the top five measures are consistent and similar, 
suggesting that the composite measure is general.

\begin{table}
\begin{tabular}{ |c|c|c|c|c|c|c|}
\hline
 0 & Full Dataset & MIS8 & LAP1 & LAP2 & STA3 & GRA5\\
\hline
 1 & Dense Textures & MIS8 & LAP1 & LAP2 & HFN & STA3\\
\hline
 2 & High Defocus & MIS8 & LAP1 & LAP2 & STA3 & GRA5\\
\hline
 3 & Outdoor Illumination & MIS8 & LAP1 & LAP2 & STA3 & GRA5\\
\hline
 4 & Random subset (50) & MIS8 & LAP1 & LAP2 & STA3 & GRA5\\
\hline
\end{tabular}\\
\caption{The top-five measures in the cfm show very little change on using
different subsets of the focal stack dataset.}
\label{nochangeTable}
\end{table}

We refer the reader to our previous work \cite{cfm} where the a similar 
composite focus measure is used for depth-from-focus. While we proposed an empirical
method to compute the composite focus measure in \cite{cfm}, in this work, we have
described a more structured approach to cluster correlated focus measures across a larger dataset
of focal stacks. The overall similarity in the composite measure suggests that is quite general.  
In this work, we use the composite focus measure to estimate the in-focus pixels within a focal stack.

\section{Focus Representation}
The composite focus measure can be used to build a compact model for focal stacks. 
We propose a representation for focal stacks that consists of the in-focus intensity
for each pixel, a secondary dual focus intensity wherever applicable, bokeh scaling of the intensity 
value wherever applicable, and the defocus kernel at the pixel at all slices of the 
captured focal stack. This representation of focus is compact compared to the full focal stack
and encodes all high-level and fine characteristics of visible scene points.

\subsection{In-focus Pixels}
The composite focus measure applied at each pixel of a focal stack provides
an estimate of the amount of focus of the pixel in each slice. In general,
the response of a focus measure across a focal volume is most reliable and 
informative at high gradients. The reliable in-focus location  
at high-gradient pixels can be propagated to other pixels
based on image content. We follow a similar principle to identify the in-focus
position for all pixels. 

\begin{figure*}
\centering
\includegraphics[width=.9\textwidth]{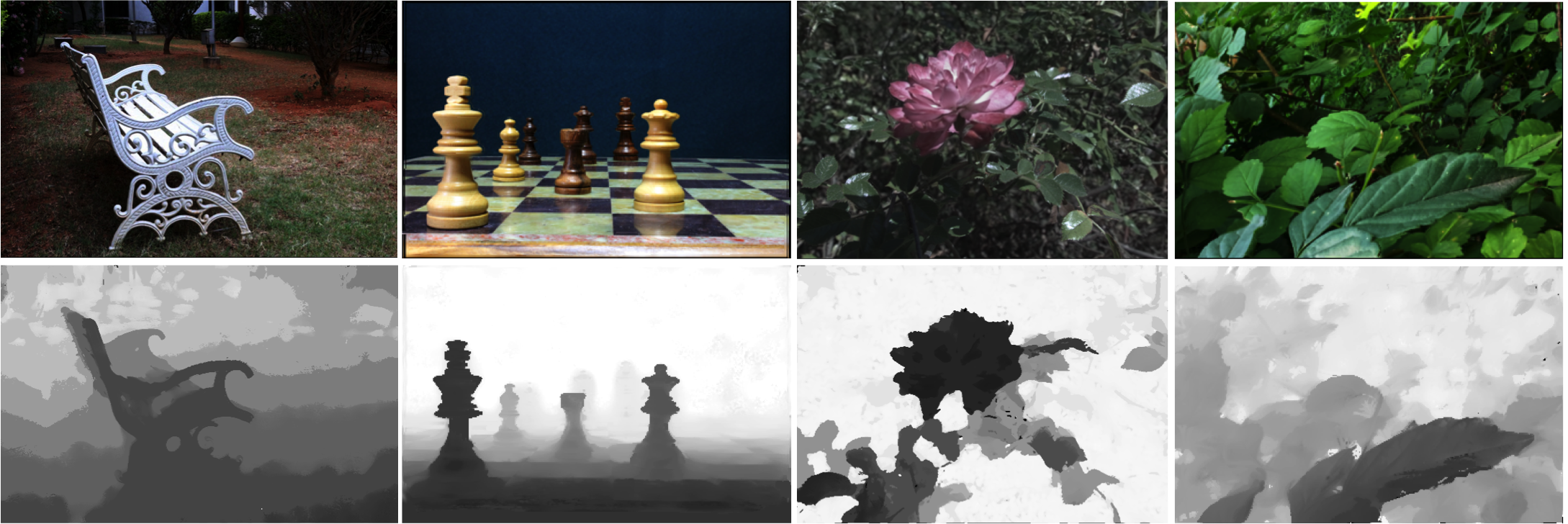}
\caption{Left-to-right: In-focus image $F_{{\cal I}}$ and focus map ${\cal I}$ for focal stacks 
captured using Canon EOS 70D, Canon EOS 1100D, Lytro Illum and the Apple iPhone 6.}
\label{Iexample}
\end{figure*}

We use cost-volume filtering \cite{CostVolume} to generate
a smooth focus map $\cal{I}$ which encodes the slice at which each pixel exhibits a
focus peak. A cost volume, representing the cost
of labeling a pixel to a focus location, is created with $k$ nodes for each pixel,
where $k$ denotes the number of slices in stack. 
The cost of labeling a pixel $p$ to a location $l$ is computed as:
\begin{equation}
C^v_l(p) = cFM(p,l)^{-1},
\label{cvf}
\end{equation}
where, $cFM(p,L)$ is the response of the cFM
evaluated at pixel $p$ in slice $l$.
The cost at each node is inversely proportional to the response of the 
cFM. Edge-aware filtering \cite{guidedFilter} of the cost volume based on a guidance image
can be used to propagate confident in-focus labels to all pixels. 
We generate a guidance image by choosing the focal slice for which the cFM 
is maximized at each pixel. This is a neighborhood-agnostic image and provides a
coarse but useful approximation of depth-edges. The final focus map
${\cal I}$ can be computed as the location of minimum cost for each pixel $p$ in the filtered
cost volume $C^{v'}$.
\begin{equation}
{\cal I}(p) = \arg\min_{i=1}^k C^{v'}_i(p)
\label{focus map}
\end{equation}
Figure \ref{Iexample} illustrates a few examples of focus maps computed using the composite
focus measure and cost volume filtering. 

\subsection{Dual-focus Pixels}
The focus map ${\cal I}$ represents the location of best focus for
each pixel.  This may not be the only focus distance
at which the pixel is in-focus. In some situations, focusing on the background leads to
 foreground objects being defocused
by a large extent such that the background objects become visible through them.
The pixels at the edges of the foreground object will now be in-focus at two focal slices,
both of which must be considered for accurate geometric modeling of the scene. In principle,
it is possible that a single pixel has more than two in-focus slice candidates. However,
this situation is impractical given the exponential increase in size of the depth-of-field with
increasing depth.

We identify pixels with dual-focus locations using the composite focus measure. This 
is a crucial attribute of wide-aperture images which is not considered by previous methods
dealing with focal stacks.
The result of cost volume filtering is a cost vector which can be reciprocated to build a focus vector
for each pixel in the scene. The maxima of this vector indicates the location of best focus and
is already encoded in ${\cal I}$. We estimate secondary peaks by applying
non-maximum suppression to the focus vector for each pixel across focal slices.
We seek pixels that show a secondary peak and lie close to a strong gradient
to build a dual-focus map ${\cal I}_d$.
\begin{equation}
{\cal I}_d(p) = \left\{ \begin{array}{cl}
l & \textrm{if } cFM(p,l)>cFM(p,l+\Delta l) \\
  & \Delta l \in [-w,w]\,\,\&\,\,l\neq{\cal I}(p)\,\,\&\,\,\nabla F_{\cal I}>t_{\nabla}\\
0 & \textrm{otherwise}
\end{array}\right\},
\label{dual}
\end{equation}
where the $w$ parameter encodes a neighborhood of 10$\%$ of focal slices
and $t_{\nabla}$ is a threshold on image gradient.
The pixels belonging to the dual-focus map ${\cal I}_d$ are 
considered separately for geometric refocusing as described in section VI.

\subsection{Scaling pixel intensities}
Natural scenes may consist of bright scene points corresponding to light-sources or 
specularities. Camera sensors have a limited dynamic range and very bright scene
points usually manifest as uniform saturated intensities. Such pixels do not
diminish in intensity even on blurring as their distributed intensity is still higher
than the dynamic range of the sensor. This phenomenon is responsible for \emph{bokeh} 
circles in wide-aperture images shown in Figure \ref{showbokeh}.

Pixel that contribute to bokeh circles need to be identified to simulate accurate geometric refocusing of 
the scene. Drawing inspiration from existing methods such as \cite{RDF,Seitz,Hach}, we identify
bokeh-causing pixels as those which have an intensity larger than a threshold $t_{\cal B}$ and do
not change their intensity across focal slices:
\begin{equation}
{\cal B}(p) = \left\{ \begin{array}{cl}
1 & \textrm{if } {\cal G}^i(p)>t_{\cal B} \,\,\,\forall i \in [1,k]\\
0 & \textrm{otherwise}\\
\end{array}\right\}.
\label{bokehEquation}
\end{equation}
The focal slice at which the bokeh
circle surrounding these points is the smallest is identified as the in-focus slice
for the bokeh scene point. All pixels that exhibit bokeh are labeled using a bokeh mask 
${\cal B}$. The true intensity of such pixels is identified by scaling up and fitting
the appropriate pillbox function that results in the largest bokeh circle around the pixel
in the focal stack. These pixels are considered separately for geometric refocusing described in 
section VI.

The in-focus map ${\cal I}$, the dual-focus map ${\cal I}_d$ and the bokeh map
${\cal B}$ form the in-focus representation for scene points. 
Additionally, we compute $F_{{\cal I}}$
which is a collection of in-focus pixels chosen from their
in-focus slice ${\cal I}$. We also compute $F_{{\cal I}_d}$ which is a sparse collection 
of secondary in-focus intensities for pixels that show dual-focus positions.
These maps and images efficiently represent the in-focus content of a 
focal stack.

\subsection{Defocus Kernels}
To estimate the defocus kernels of different pixels,
we use the in-focus intensity $F_{{\cal I}}(p)$ of a pixel $p$ 
as a representation of the luminosity of its scene point. 
Using this luminosity, we estimate the degree of
defocus or the blur-radius for that pixel at other focus positions.
Defocused intensities of 
the points from different depths may contribute to the same pixel on the sensor. 
We therefore do not use pixels that lie close to depth edges and restrict kernel
estimation to equi-focal pixels. 

We isolate the pixels that belong to similar regions in depth based on
${\cal I}$. These pixels come from scene points that were close to the same focal plane 
in the world and are called equi-focal pixels. They appear equally focused
(or equally defocused) in any focal slice.
The defocus radii of equi-focal pixels can be calibrated in a spatially-invariant manner.
The defocus radii of pixels close to a depth-edge can consequently be computed from their
nearest equi-focal region.

We use a generative approach to model the defocus kernel that an in-focus pixel 
subtends at other focal slices. For a defocused scene point,
the rays from the point spread out across a defocus kernel of an appropriate size
and shape (Figure \ref{GeomRadii}). This can be modeled as an
intensity distribution operation from the source pixel to several pixels.
Distribution of intensities from a source point is analogous to kernel-splatting
methods \cite{KerSplat} where each point may have a different contribution. 
The intensity distributed to a pixel $p$ from its neighbors $q$ can be modeled as:
\begin{equation}
\Delta I_q(p) = h(p,q)\circledast F_{{\cal I}}(q) \hspace{3ex} \forall q \in N_p,
\label{pushEnergyEquifocal}
\end{equation}
where $\Delta I_q(p)$ is the intensity that pixel $q$ distributes to pixel 
$p$, $F_{{\cal I}}(q)$ the intensity of $q$, $N_{p}$ the neighborhood
around $p$, $h(x,y)$ the unknown defocus kernel or point-spread-function (PSF)
that we would like to model and $\circledast$ denotes convolution. $h$ is typically 
modeled as a normalized 2D Gaussian kernel with uniform sigma in both axes.

If the above model is restricted to equi-focal pixels, the defocus kernel 
can be estimated by finding the best spatially invariant convolution kernel
that converts a source pixel's intensity into the target pixel as
\begin{equation}
I(p) = \sum_{q \in N_{p}} \frac{1}{\sigma \sqrt{2\pi}}\mathrm{e}^{(\frac{-d^2}{2\sigma^2})}F_{{\cal I}}(q),
\label{gaussEquifocal}
\end{equation}
where, $\sigma$ denotes the space-invariant defocus radius, $d$ the Euclidean distance
between pixels $p$ and $q$ and $g(q)$ the focused intensity for each pixel in $N_{p}$.
Here $I(p)$ is the rendered intensity of a defocused pixel $p$ and can be compared to 
different focal slices in ${\cal G}$ to estimate corresponding defocus kernels.

\begin{figure}
\centering
\includegraphics[width=.35\textwidth]{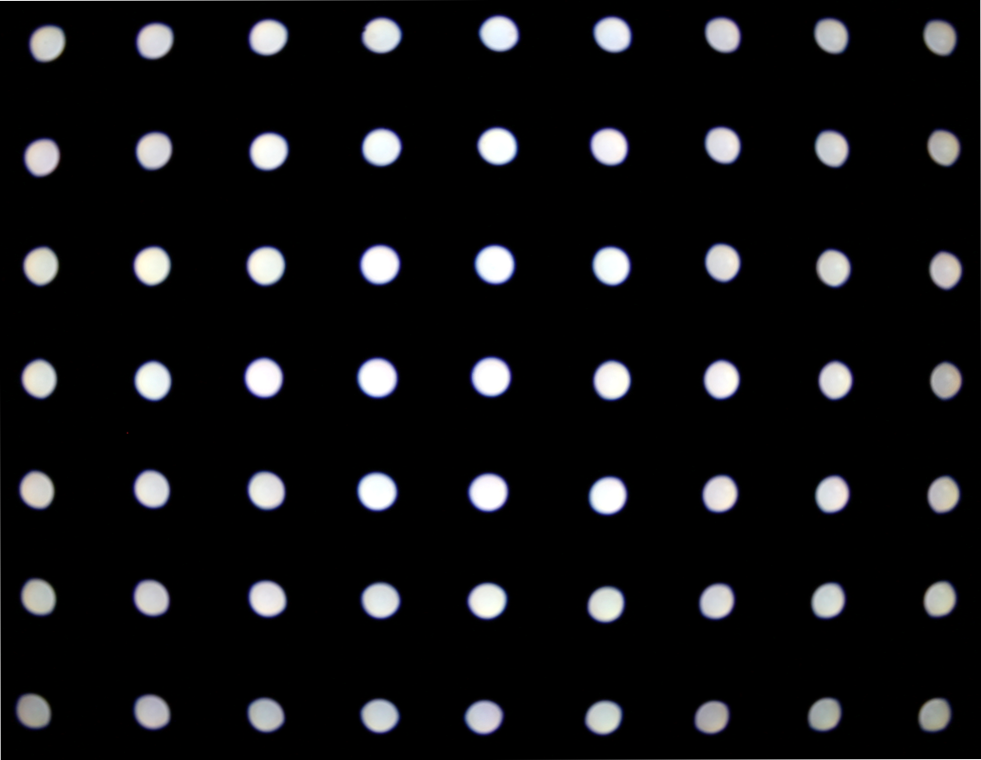}
\caption{The change in shape of defocus kernels due to vignetting on a Canon 1100D DSLR at f/3.5 with a focal length of 18mm.}
\label{psfShapeChange}
\end{figure}

We use geometric calibration to compute the defocus kernels by iterating across equi-focal
pixels in a piece-wise manner. 
We estimate the size and the variable shape of defocus kernels similar to approaches such
as \cite{Kee, Shih, Hach}. The $\sigma$ of the blur kernel depends on 
the distance between the current focus position and the pixel's in-focus position 
and the shape of the kernel depends on the location of the pixel on the sensor. Pixels
towards the edges exhibit vignetting (clipping) and thus result in shortened blur kernels.
Figure \ref{psfShapeChange} shows a defocused image of multiple defocused small
point light sources captured under low-light conditions. The defocus shapes at the extremeties 
are visibly shortened compared to those towards the center. 

\begin{algorithm}
\caption{Size and Shape of Defocus Kernels}\label{blurCompare}
\begin{algorithmic}[1]
\Procedure{$\Pi$}{$x,y,{\cal I}(x,y),T$}
\State $s_{xy} \gets |{\cal I}(x,y)-T|$ 
\State min $\gets \inf$
\State $p_{mn} \gets  F_{\cal I}(x,y,[m,n])$
\State $r_{mn} \gets  {\cal G}_T(x,y,[m,n])$
\While{$z$ in sizes($s_{xy}$)}
\While{$h$ in shapes($x,y$)}
\State $\hat{p}_{mn} \gets gaussianBlur(p_{mn},z,h)$
\State $d \gets \sum\sum |\hat{p}_{mn}-r_{mn}|$
\If{$d<min$}
\State $\Pi({\cal I}(x,y),T) \gets (z,h)$
\State min $\gets d$
\EndIf
\EndWhile
\EndWhile\label{euclidendwhile}
\State \textbf{return}
\EndProcedure
\end{algorithmic}
\end{algorithm}

We estimate blur kernels between all pairs of focal
slices using a blur-and-compare framework. For each pair, we use the in-focus
pixels $F_{\cal I}$ and compute their largest continuous rectangular region
as a reference sub-image. The centre of the rectangular region
and its separation from the center of the image is noted to account for vignetting. 
We then blur the in-focus pixels from one slice with several candidate
size and shape parameters and compare the blurred images with the other slice
to compute the best match. The shapes are chosen from a set of shapes motivated by \cite{Hach}
and shown in Figure \ref{psfShapeChange}. The defocus kernel
between a pair of focal slices is computed in a bidirectional manner and the
size and shape parameters are recorded for both directions. Algorithm \ref{blurCompare} outlines
the process of estimating the blur radii between a pair of focal slices.

In Algorithm \ref{blurCompare}, $p_{mn}$ is a patch of size $[m,n]$ close to pixel $(x,y)$ in $F_{\cal I}$ such that all the
pixels in the patch have the same value of ${\cal I}$ and $r_{mn}$ is the corresponding patch from 
the target focal slice ${\cal G}_T$. The collection of tuples $\Pi$ encodes the size and shape of defocus kernels
between all pairs of focal slices in a focal stack. Since this is explicitly calibrated for each stack, 
minor differences in camera design are automatically encoded. $\Pi$ is a collection of at-most 
$k^2$ sizes and $k^2$ shapes and thus represents the defocus content within the focal stack in 
a compact manner.

We can now represent a focal stack using the ${\cal I}, F, {\cal B}$ and $\Pi$ constructs. 
This is a compact and robust representation that encodes all the focus and defocus properties of the
scene points. The ${\cal I}$ maps encode the indices at which the pixel is likely to be
in-focus while the $F$ images store photometrically accurate intensities corresponding
to these indices, with ${\cal B}$ labeling the pixels that require intensity scaling. 
The $\Pi$ map encodes the size and shape of defocus that the pixel exhibits in all other 
not-in-focus slices. The overall storage complexity of our representation is two dense images ${\cal I}$ (single channel)
and $F_{\cal I}$ (RGB),
three sparse images ${\cal I}_d$ (single channel), $F_{{\cal I}_d}$ (RGB) and ${\cal B}$ (single channel) and at most $k^2$ 
defocus sizes and shapes. A full focal stack on the other hand stores $k$ RGB images with no implicit information 
about scene content. Our representation is ideally suited for geometrically accurate and photo-realistic scene refocusing,
described in the next section.

\section{Geometric Scene Refocusing}
We describe a geometric approach to scene refocusing using our representation for focal stacks.
Our approach pays explicit attention to fine effects such foreground-background occlusions,
mixing of blur kernels at depth-edges, vignetting and bokeh circles. 
Refocusing is formulated as the task of rendering a target depth-of-field,
where the limits of the depth-of-field are defined by focal slice indices $T$
chosen from the unique labels in the focus map ${\cal I}$. 

Generating a novel focused image from our representation is a two-step process.
The first step is to identify the appropriate intensity for each pixel and the second step is
to distribute this intensity to its neighbors based on the defocus kernel at the pixel. 
We synthesize a refocused image in a piece-wise manner. For all unique labels in the
focus map ${\cal I}$, the intensity of the corresponding pixels (or dual-pixels)
is considered as the radiance of their scene points. This radiance is scaled 
appropriately for pixels that also belong to the Bokeh mask ${\cal B}$. 
The radiance of these pixels is then distributed to its neighbors based
on the defocus kernel that pixel subtends on the target focus
position. This follows the distributive energy model for wide-aperture images
suggested in Equation \ref{pushEnergyEquifocal}. The pixels in
a scene may correspond to different depths and the spatially varying defocus kernels
need to be applied in the right order according to the distribution model:
\begin{equation}
\Delta I_q(p) = \sum_{q \in N_{p}} H(p,q,\sigma_q).F_{\cal I}(q),
\label{gaussMultiDepthNorm}
\end{equation}
where $H(p,q,\sigma_q) = \frac{1}{\sigma_{q} \sqrt{2R}}\mathrm{e}^{(\frac{-d^2}{2\sigma_{q}^2})}$
is a Gaussian point spread function with an appropriate $\sigma_q$ and corresponding shape.

\begin{figure}
\centering
\includegraphics[width=.35\textwidth]{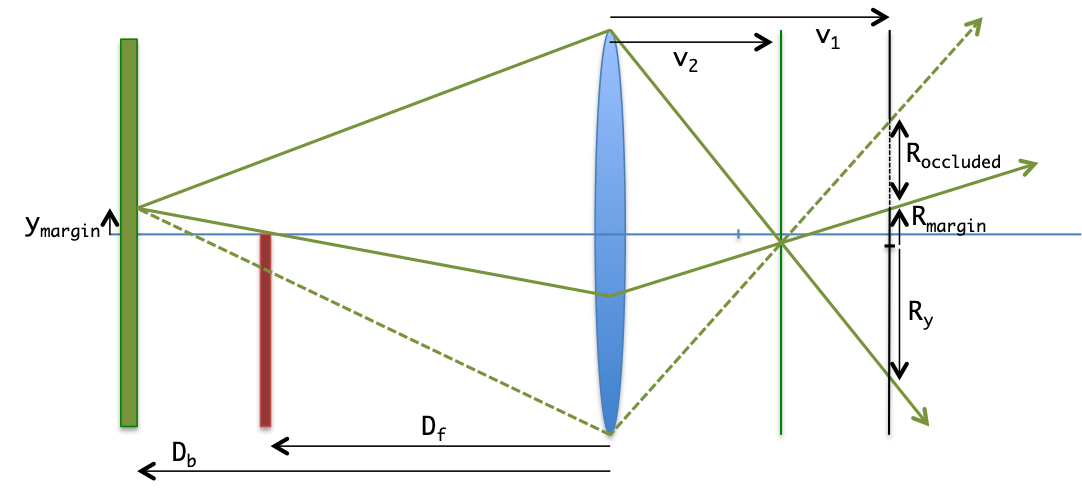}
\caption{The presence of the foreground partially occludes the background kernel by an amount $R_{occluded}$. The 
occlusion free component $R_{margin}$ can be computed using $D_f$, $D_b$, $R_y$, $A$ and $y_{margin}$ as shown in
Eqn. \ref{R_margin}}
\label{occlusionRange}
\end{figure}

The order of energy distribution becomes important close to depth-edges.
As shown in Figure \ref{occlusionRange}, when the sensor is placed at a position $v_1$
which is a near-focus position, the defocused contribution of the background pixels is partially
occluded by the presence of the foreground object. Note that our representation
consists of dual-focus pixels and captures these pixels behind visible foreground segments.
The energy distributed from a background pixel should not freely merge with foreground pixels
irrespective of camera and scene geometry. To model the situation shown in Figure \ref{occlusionRange},
we evaluate the impact of partial occlusions in a geometrically accurate manner.

\begin{figure}
\centering
\includegraphics[width=.35\textwidth]{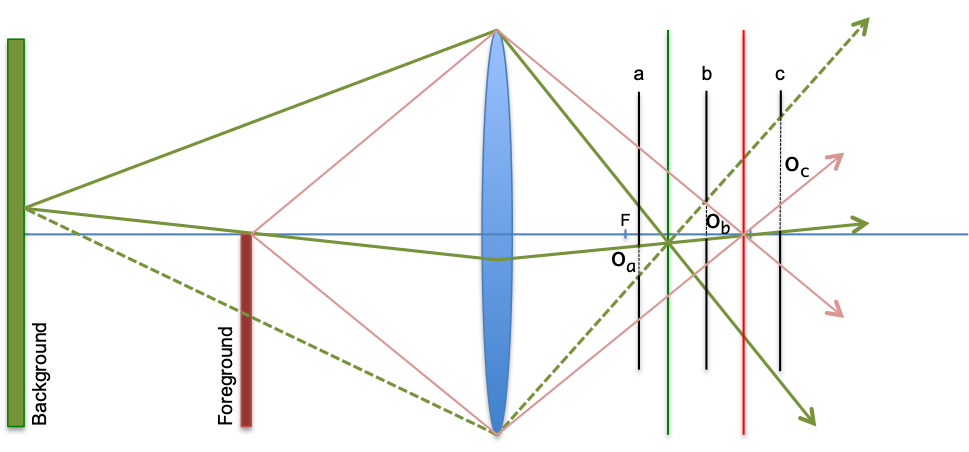}
\caption{The defocused contribution of background pixels is occluded by the foreground. At sensor positions
$a$, $b$ and $c$, the occlusion due to the foreground on the background pixel's contribution is indicated by the
dotted portion $o_a$, $o_b$ and $o_c$ respectively.}
\label{occlusion}
\end{figure}

The amount of partial occlusion of the defocus kernel depends on camera and scene geometry while the 
shape of the occlusion depends on the shape of the depth-edge. The size of the occlusion varies 
from 0 to 100\% from a limiting point above the principal axis to a symmetric point below the principal
axis in Figure \ref{occlusionRange}.
Figure \ref{occlusion} illustrates three regions of focus that
are of importance, one focused beyond the background $a$, the second focused
between the foreground and the background $b$ and the third focused closer than the
foreground $c$. The occluded portion of defocus kernels is denoted by $o_a$, $o_b$ and $o_c$.

\begin{itemize}
\item At a sensor position close to $a$, the foreground pixels are severly defocused 
and dominate the image content. A small occlusion $o_a$ in the defocus kernel of the 
background has minimal impact on image content.
\item At sensor positions close to $b$, the spread of the background pixels overlaps with 
slightly defocused foreground, however, no background pixel overlaps with in-focus side of 
foreground pixels (above the principle axis in the diagram).
Thus the contribution of a defocused background pixel to a foreground pixel must be 
disallowed based on the size and shape of $o_b$.
\item When the sensor plane is focused in front of the foreground such as in position $c$, 
the background pixel spread overlaps above the principal axis and needs to
be shortened based on $o_c$. 
\end{itemize}

To model partial occlusions in our refocusing algorithm,
we introduce an occlusion coefficient $\beta$ which geometrically restricts the 
contribution of a source pixel to a 
target pixel. The energy distribution model for a pixel is reformulated as

\begin{equation}
\Delta I_q(p) = \sum_{q \in N_{p}} \beta_{qp}H(p,q,\sigma_q).F_{\cal I}(q).
\label{gaussBeta}
\end{equation}

For background pixels below the occlusion boundary, $F_{{\cal I}_d}(q)$ may be used wherever
applicable. The occlusion co-efficient is only considered when background-to-foreground
mixing occurs, i.e. when ${\cal I}_q-{\cal I}_p>t_{\beta}$, with $t_{\beta}$ indicating a threshold
on number of focal slices.
When the precise geometry of the lens and scene is known, the $\beta_{qp}$ parameter can be 
computed based on the distance between pixels $p$ and $q$ and $R_{margin}$ \cite{Subhashis2001}, where,

\begin{equation}
R_{margin} = \frac{R_{q}}{A/2} . \frac{D_f}{D_b-D_f} . y_{margin}.
\label{R_margin}
\end{equation}
$D_f$ and $D_b$ denote 
the object-side depth for the foreground and background objects, $y_{margin}$ is the distance of source
point $y$ from the depth edge on the object side and $R_{q}$ is the geometric kernel radius
for pixel $q$ for the current sensor position. The metric units of 
$R_{margin}$ and $y_{margin}$ can be converted to pixel units. In a focal stack,
absolute depth values for source and target pixels $q$ and $p$ may not be known but their relative depth
values are indicated by the focus map ${\cal I}$ and the depth ratio can be computed as
$\frac{{\cal I}(p)}{{\cal I}(q)-{\cal I}(p)}$. The $y_{margin}$ parameter can be estimated 
up to an unknown scale factor by calculating the distance of the source pixel from
the target pixel in the ${\cal I}$ image. This can be converted to metric units
using an approximate estimate of closest and farthest focus positions. 
Using the Euclidean distance $d$ between pixels $p$ and $q$, the value of
$y_{margin}$ can be computed as $y_{margin} = \frac{k{\cal I}(q)+c}{f}\,d,$
where $k$ and $c$ are approximate depths of the nearest and
farthest focus positions and $f$ is the focal length of the lens. These constants
can be estimated exactly if the capture-time sensor positions are known (using APIs such
as MagicLantern). For free-form focal stacks, the constants can
be approximated based on the visible scene content. Using these constructs,
the occlusion coefficient $\beta_{qp}$ is defined as:
\begin{equation}
\beta_{qp} = \left\{ \begin{array}{cl}
1 & \textrm{if } d \le R_{margin}\\
0 & \textrm{otherwise}
\end{array}\right\}.
\label{beta}
\end{equation}
It may be noted that for dual focus pixels, the sign of $y_{margin}$ is negative and the 
direction of intensity distribution from the dual pixel is automatically inverted.

\begin{figure*}
\centering
\includegraphics[width=.95\textwidth]{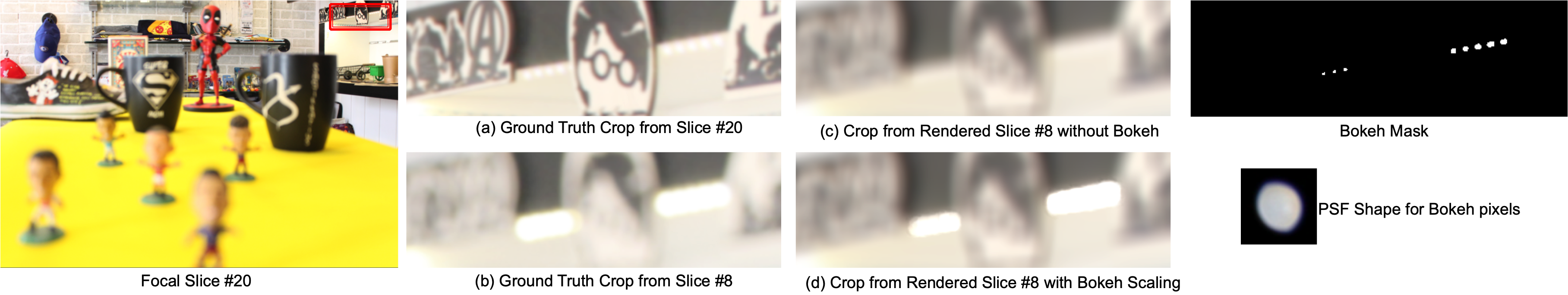}
\caption{Bokeh simulation using saturated intensities and our geometric refocusing algorithm. 
The bokeh pixels identified in slice \#20 are shown in the Bokeh mask. The asymmetric shape of the 
point-spread-function is selected from Figure \ref{psfShapeChange} corresponding to the spatial location
of the bokeh pixels. The defocused version of figure a using our algorithm without bokeh scaling is shown in figure c (PSNR: 23.5dB, SSIM: 0.90) and with
our bokeh scaling approach is shown in figure d (PSNR 28.1dB, SSIM:0.95). Note that our algorithm better simulates intensity saturation 
visible in the ground-truth figure b.}
\label{showbokeh}
\end{figure*}

\begin{figure}
\centering
\includegraphics[width=.48\textwidth]{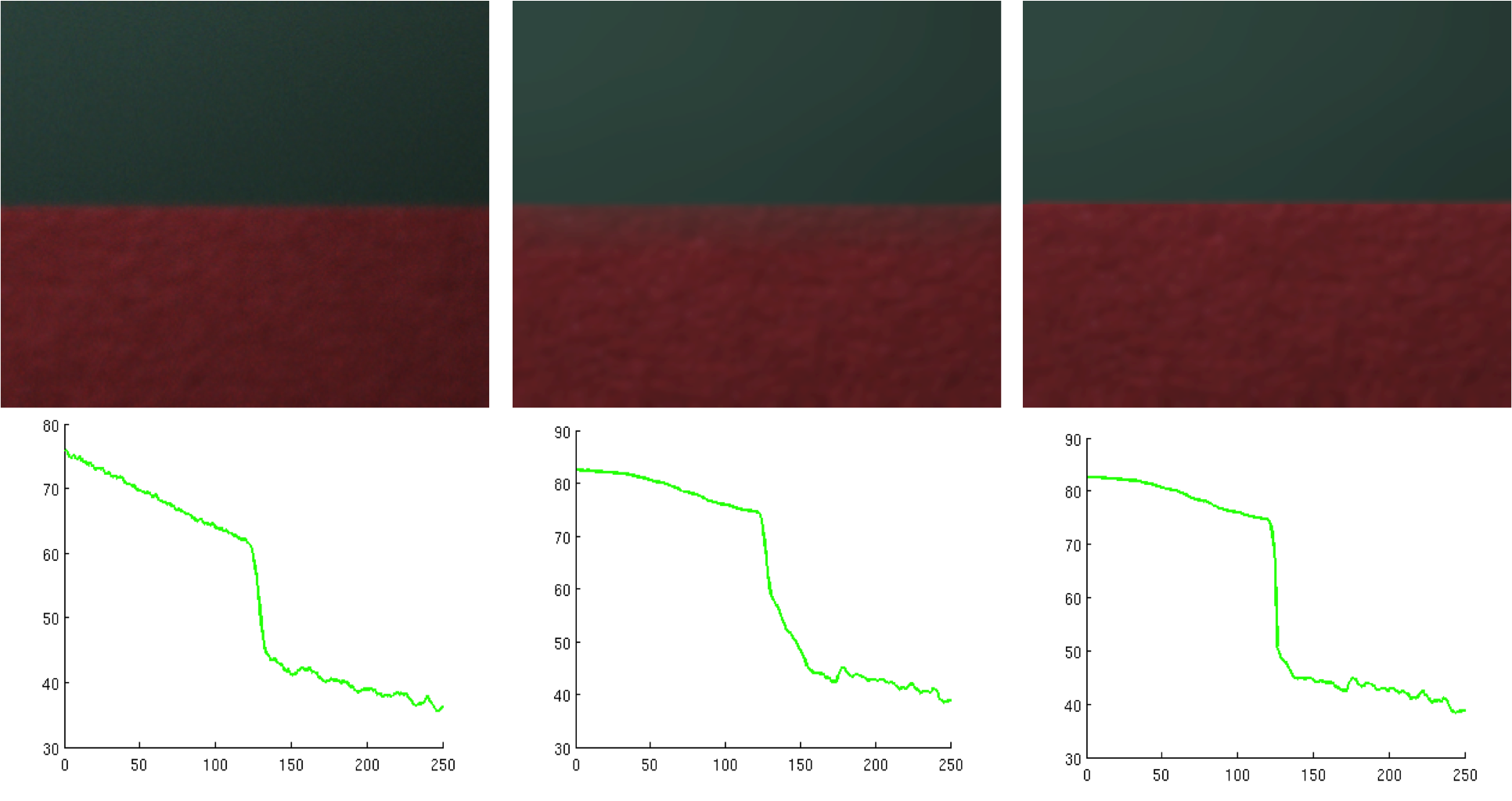}
\caption{A synthetic red-green depth-edge experiment. The target focus position is similar to $o_c$ from 
Figure \ref{occlusionRange}. From left-to-right: Ground truth image, reconstructed image without
$\beta$ and reconstructed image with $\beta$.
The second row show the corresponding plots of green
intensity from top-to-bottom in the images. Our model with $\beta$ correctly 
restricts distribution from background to foreground pixels.}
\label{RedGreen}
\end{figure}

Algorithm \ref{refocusAlgo} outlines our overall method for rendering novel focused positions from
our representation of a focal stack. In a front-to-back manner, all the focus
labels in ${\cal I}$ are considered. A target pixel is generated by accumulating intensities
distributed by its neighboring pixels. Bokeh scaling, dual-pixel intensities, partial
occlusions at depth-edges and kernel shortening are considered explicitly before intensity distribution.

\begin{algorithm}
\caption{Refocus Scene to Focus Position(s) T}\label{refocusAlgo}
\begin{algorithmic}[1]
\Procedure{refocus}{${\cal I},{\cal I}_d,\Pi,T$}
\For{all labels $l$ in ${\cal I}$}
\For{all pixels $q$ with ${\cal I}(q)=l$ or ${\cal I}_d(q)=l$ }
\State $i_p \gets F_{\cal I}(q)$ or $F_{{\cal I}_d}(q)$
\If{${\cal B}(q)=1$}
\State $i_q \gets$ scaleBokeh($i_q$)
\EndIf
\For{all pixels $p$ neighboring $q$}
\State kernel $\gets \Pi({\cal I}(q),{\cal I}(p))$
\State $i_p \gets i_p+$ applyKernel($i_q,\beta_{qp},$kernel);
\State
\EndFor
\EndFor
\EndFor
\State \textbf{return}
\EndProcedure
\end{algorithmic}
\end{algorithm}

\section{Experiments \& Results}
We perform quantitative and qualitative evaluation of our representation in terms of its capability of 
reconstructing the focal stack and producing novel refocused renderings of
the scene.
We also compare our method with that of other state-of-the-art methods in post-capture
focus control and manipulation. 

\subsection{Dataset}
\begin{figure*}
\centering
\includegraphics[width=.98\textwidth]{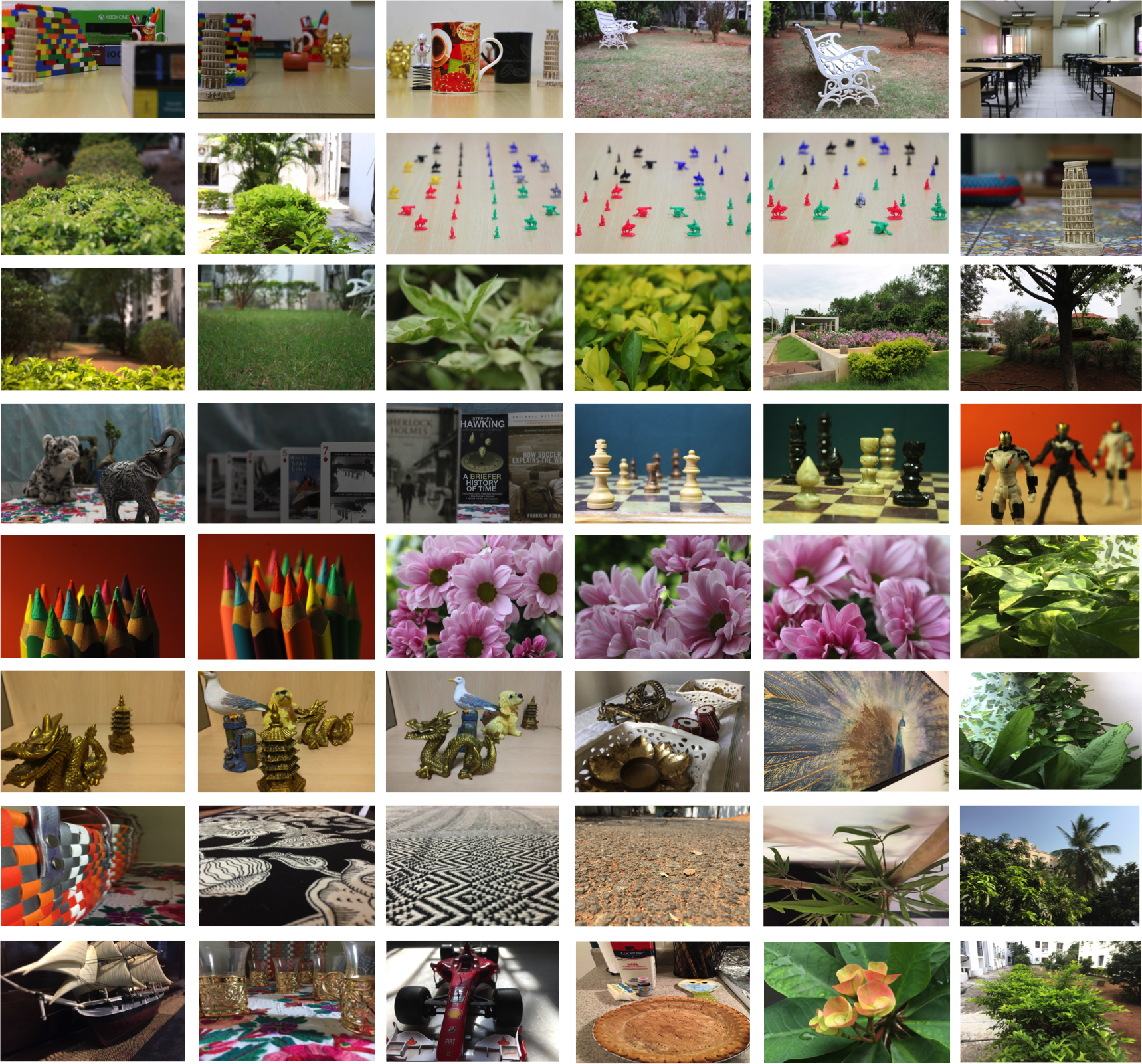}
\caption{An illustration of the different types of scenes in our focal stack dataset captured using DSLR 
and mobile cameras.}
\label{dataset}
\end{figure*}

We use focal stack dataset of about 320 focal stacks. This dataset is composed of 100 focal 
stacks from the light-field saliency dataset \cite{LFSD}, 80 sparsely sampled
stationary focal stacks from the Autofocus dataset \cite{eccvBrown}, 20 focal stacks
that are available publicly across different works \cite{Boshtayeva} and  
120 focal stacks that we captured using DSLR and mobile cameras. Our DSLR focal 
stacks are captured using Magic Lantern on a Canon EOS 70D and a Canon EOS 1100D.
Our mobile focal stacks are captured using an iPhone 6 and an HTC One X (Figure \ref{dataset}).
Each focal stack is rescaled to the order of 1M pixels while preserving the aspect
ratio. All our experiments are performed on RAW images (or on images rescaled by an appropriate gamma 
factor). To eliminate pixel misalignment due to magnification, we use the enhanced correlation
coefficient maximization approach \cite{ECC}.  The thresholds used in our approach are
$t_\nabla$=$20$, $t_{\cal B}$=$0.9$ and $t_\beta$=$30\%$.
To compute the composite focus measure,
the response of each focus measure is averaged across three different region of support
sizes 3$\times$3, 7$\times$7 and 11$\times$11.
The cumulative consensus response for each focus
measure is thereby computed across a corpus of about 1.2 billion pixels across the 320 focal stacks.

\subsection{Reconstructing Focal Slices}
\begin{figure*}
\centering
\includegraphics[width=.9\textwidth]{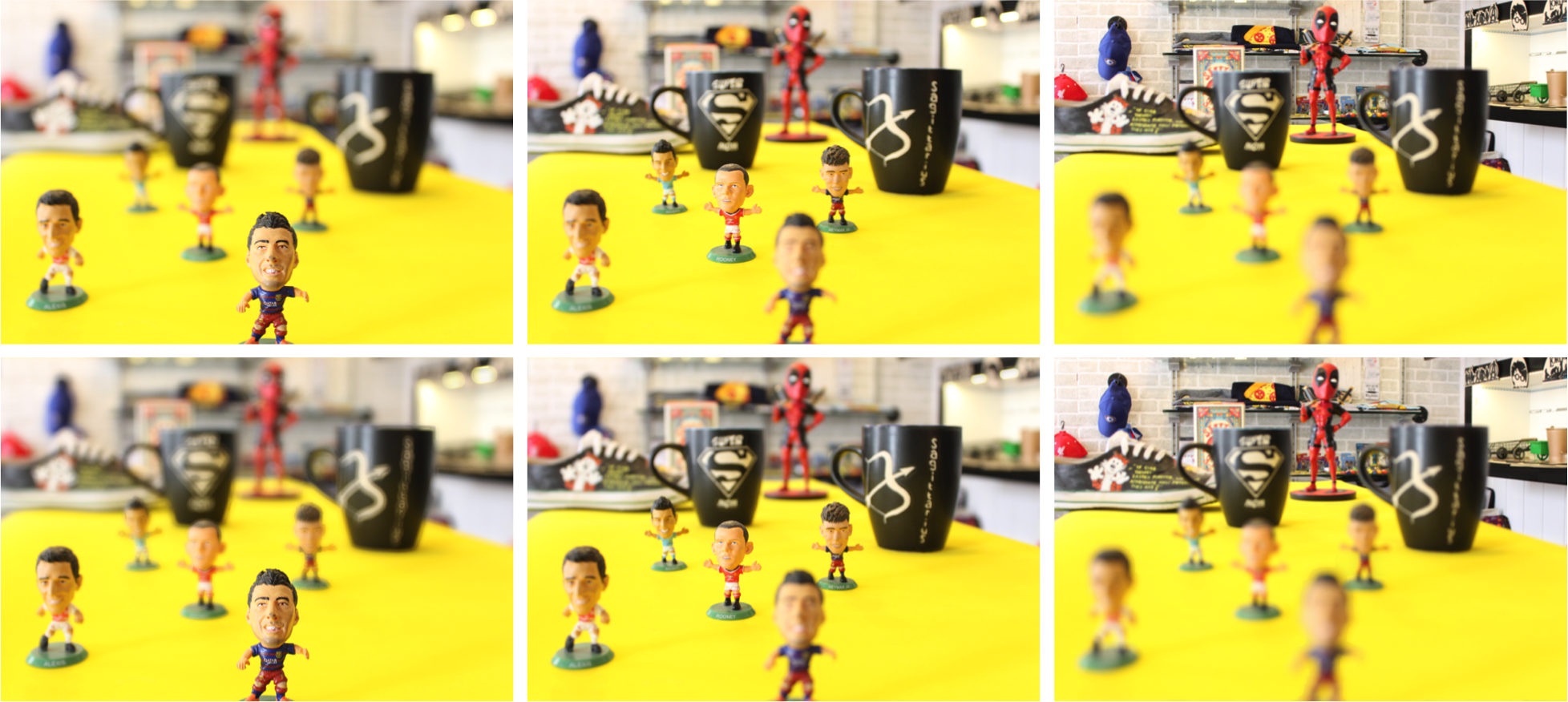}
\caption{Reconstructed Focal Slices 2, 9 and 18 for a focal stack
using our representation and refocusing algorithm. The top-row
shows the images captured using a DSLR camera and the bottom row shows the images reconstructed using our
representation and our geometric refocusing algorithm.}
\label{compareFS}
\end{figure*}

The litmus test for our representation of focus is its ability to reconstruct 
the focal slices in the stack. The model is expected to not only capture the basic 
blur profile of scene elements but to also re-create some of the fine transitions
at depth-edges. We perform a quantitative and qualitative analysis of the reconstruction
quality of our model and refocusing algorithm. Figure \ref{compareFS} illustrates an example on
a test focal stack of twenty-slices that is not a part of our focal stack dataset. 
The figure shows a comparison at three
different focus positions - slice 2, slice 9 and slice 18. In the top-row left is the image
captured using the camera and in the bottom row is the image refocused using our
representation and refocusing algorithm. On a test dataset of ten such focal stacks that were
not included in the computation of the composite focus measure, we achieve an average
reconstruction PSNR of 41.2 dB per focal slice. This score
suggests that our model for focus and refocusing algorithm works well. All images in this
paper are best viewed in color.

\begin{figure*}
\centering
\includegraphics[width=.9\textwidth]{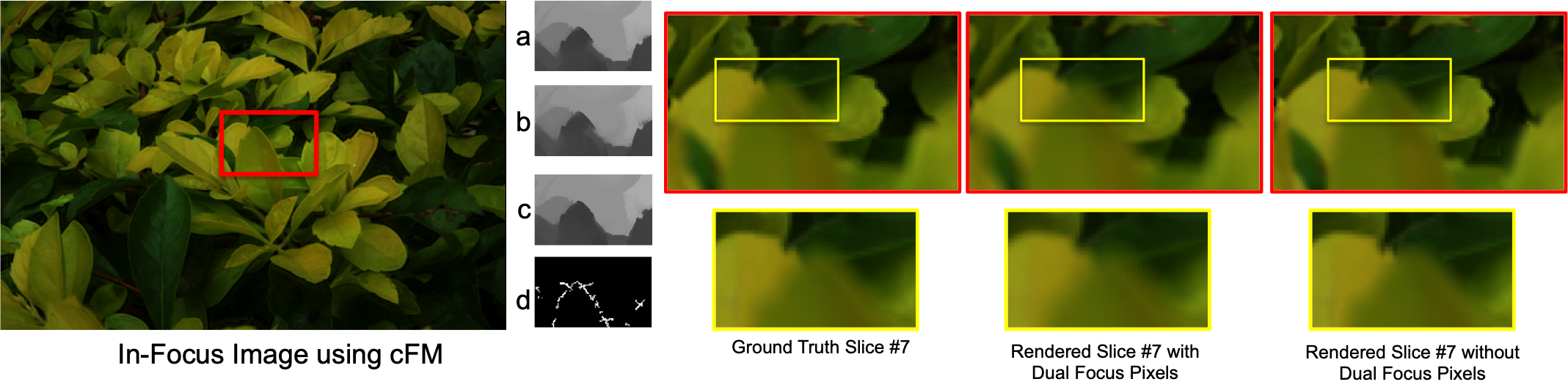}
\caption{The in-focus image shown on the left is the $F_{\cal I}$ image created using the
composite focus measure. Inset (a) shows the focus map ${\cal I}$ on using the LAP2 focus measure
alone, inset (b) shows the focus map ${\cal I}$ on using the WAV1 focus measure alone, while inset (c)
shows the focus map ${\cal I}$ on using the composite focus measure. Inset (d) is the dual focus pixel
map ${\cal I}_d$. On using the dual focus pixels, the background dark green leaf is visible 
through the blurred light green foreground, which is similar to ground truth. Ignoring the dual pixels
misses this fine effect as seen in the rightmost image.}
\label{dualPeaks}
\end{figure*}

\begin{figure*}
\centering
\begin{subfigure}
\centering
\includegraphics[width=.30\textwidth]{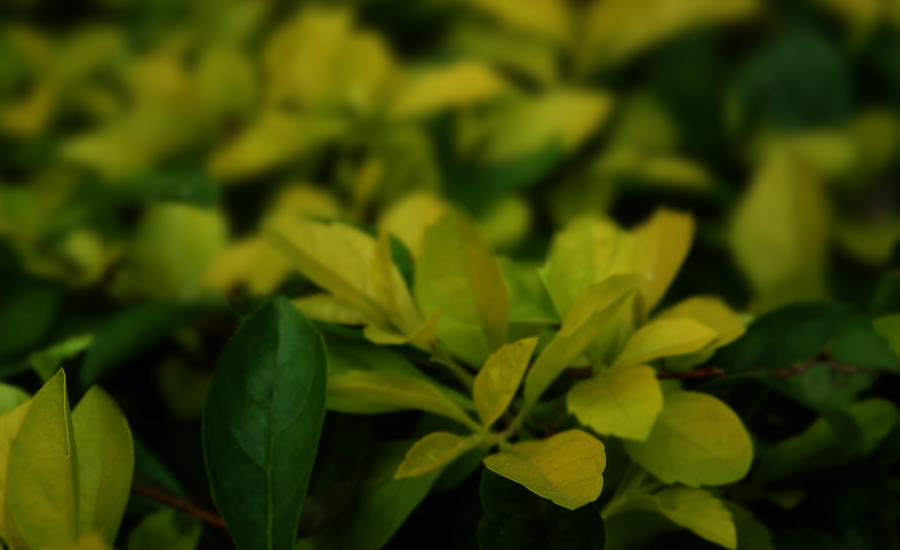}
\end{subfigure}
\begin{subfigure}
\centering
\includegraphics[width=.30\textwidth]{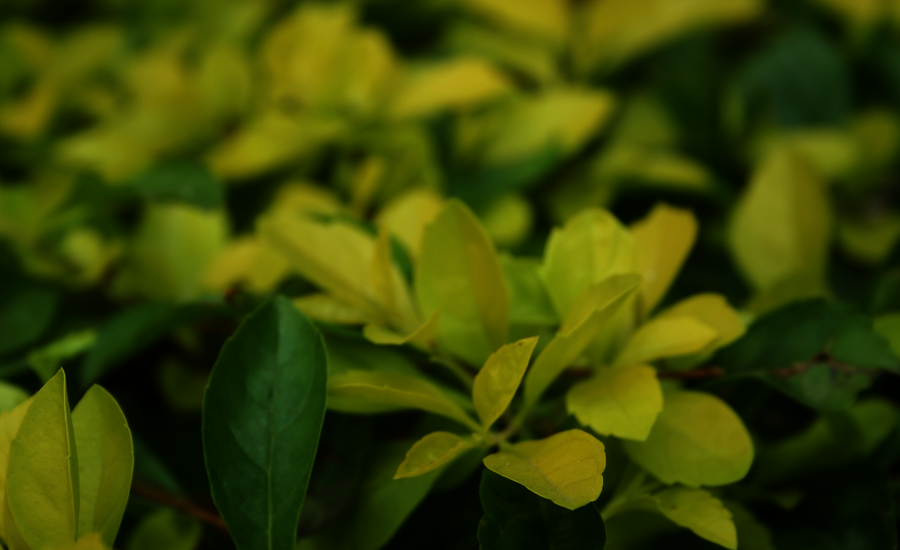}
\end{subfigure}%
\begin{subfigure}
\centering
\includegraphics[width=.30\textwidth]{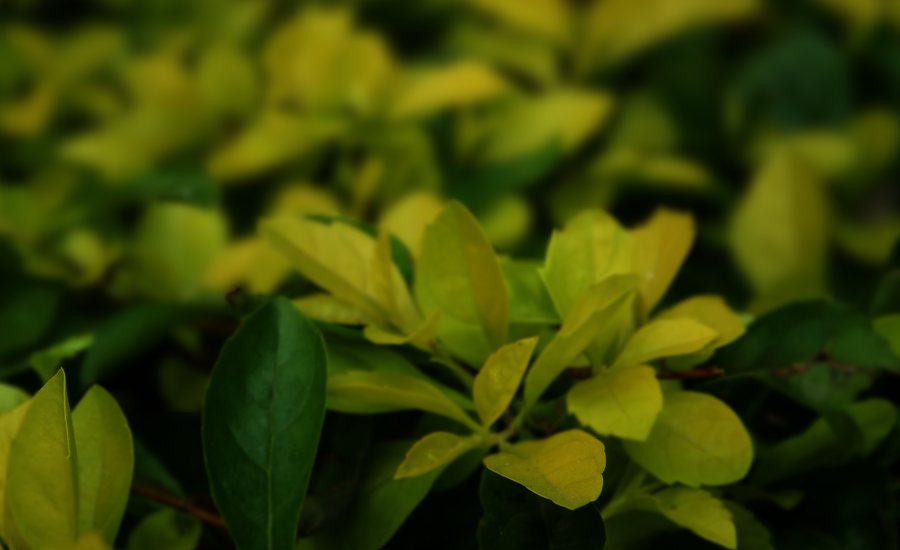}
\end{subfigure} \\[1ex]
\begin{subfigure}
\centering
\includegraphics[width=.30\textwidth]{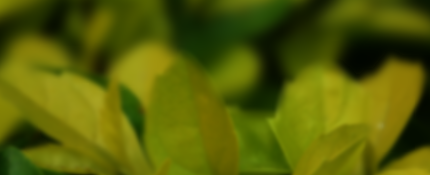}
\end{subfigure}
\begin{subfigure}
\centering
\includegraphics[width=.30\textwidth]{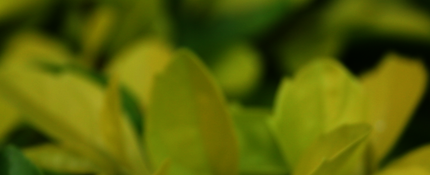}
\end{subfigure}%
\begin{subfigure}
\centering
\includegraphics[width=.30\textwidth]{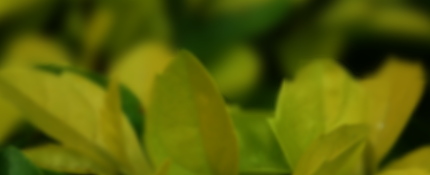}
\end{subfigure}
\caption{First row: Left to right: Simulated focal slice 2 without $\beta$ (PSNR: 34.3dB), true focal slice 2, simulated focal slice 2
using $\beta$ (PSNR 38.8dB). Second row: Insets corresponding to the first row. Notice how the background cuts into the foreground
at several leaf edges which is visible in the inset images.}
\label{betaError}
\end{figure*}

\paragraph{Ablation Study}
To elaborate the impact of the dual-focus representation, the bokeh scaling
parameter and the occlusion coefficient, we apply of our refocusing algorithm
without the three factors. Figure \ref{dualPeaks} shows an example where
our secondary focus peak estimation is necessary for simulating fine focus effects close to depth
edges. Figure \ref{showbokeh} demonstrates the benefit of using our bokeh mask for 
intensity scaling and the application of defocus kernel shortening to simulate vignetting. 
Figure \ref{betaError} illustrates the utility of the occlusion co-efficient for accurate
focus simulation at deep depth-edges. The qualitative and quantitative improvement provided 
by our approach is visible across these examples.

To further study the impact of the occlusion co-efficient at depth edges, 
we capture a focal stack of a test scene consisting of a red foreground plane and 
a green background plane at fixed distances from the camera. We reconstruct a 
target focal slice focused in front of the foreground object 
(similar to $o_c$ of Figure \ref{occlusionRange}).
We compare this image with a ground-truth focal slice.
We quantitatively study the impact of using our defocus model without the
occlusion coefficient $\beta$, in which case the green background incorrectly
defocuses into the red foreground which is visible on closely observing Fig. \ref{RedGreen}.

\begin{figure*}
\centering
\includegraphics[width=.9\textwidth]{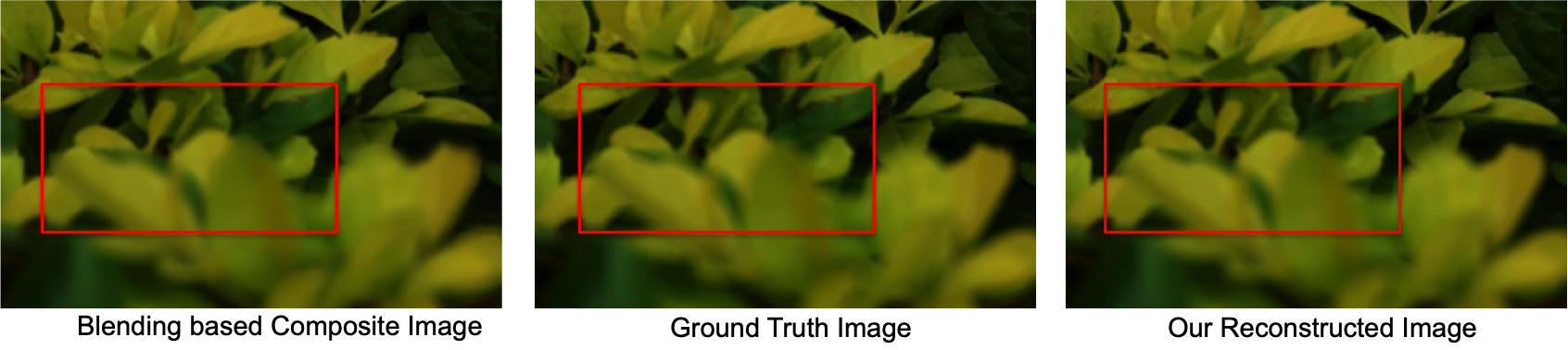}
\caption{A comparison of blending based focal slice reconstruction of \cite{barron}
with our geometric algorithm for refocusing. Note the patchy nature of depth-edges
in the blending based composite image (PSNR: 28.6dB). The dual focus pixels in our representation 
enable accurate reconstruction (PSNR: 33.7dB).}
\label{blendComp}
\end{figure*}

\begin{figure*}
\centering
\includegraphics[width=.9\textwidth]{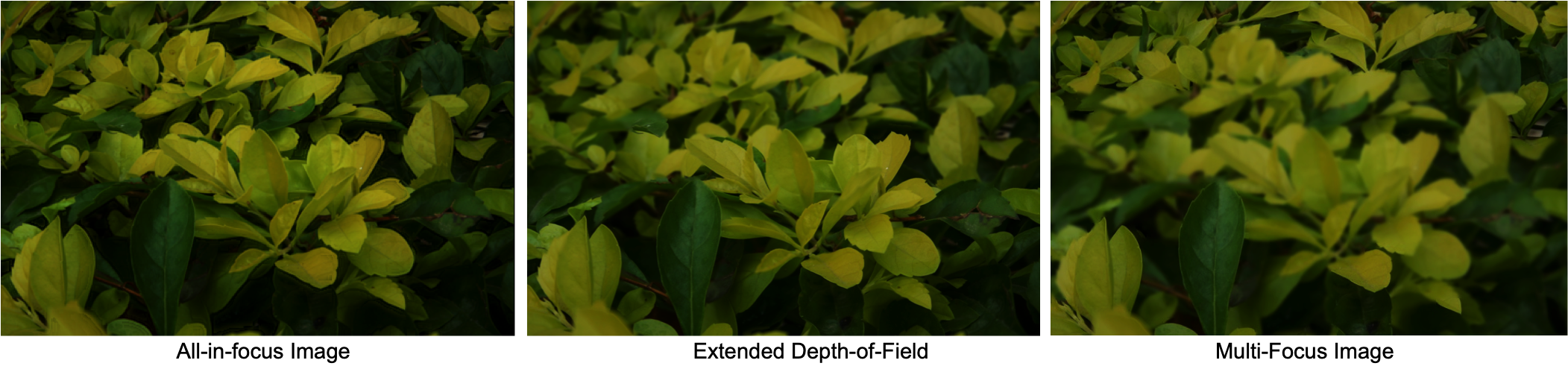}
\caption{An all-in-focus image (all focal slices in focus), an extended focus image (with multiple contiguous 
focal slices in focus) and a non-photorealistic image (with non-contiguous focal slices in focus) using our representation
and refocusing algorithm. }
\label{others}
\end{figure*}

\subsection{Refocusing the Scene}
Using our representation, we can create novel focus positions for the scene
by changing the target focus position $T$ of algorithm \ref{refocusAlgo}. 
The target position T can be a single 
focus position chosen from one of the labels in ${\cal I}$ or a
set of labels which may or may not be contiguous.

\paragraph{All-in-focus image}
In this case, $T$ consists of all labels from ${\cal I}$.
The refocusing algorithm collects the in-focus pixels from all labels. Some focal stacks
with their focus map and corresponding in-focus image are shown in Figures \ref{Iexample},\ref{others}.

\paragraph{Extended depth-of-field}
When $T$ consists of a contiguous subset of labels from ${\cal I}$, an extended depth-of-field
image can be created. The defocus kernels for such an image are chosen from the limiting labels
of $T$ on either side. An extended depth-of-field image is shown in Figure \ref{others}.

\paragraph{Non-photorealistic Focus}
When $T$ consists of multiple disjoint subsets of labels from ${\cal I}$, a non-photorealistic
rendering can be created. The defocus kernels can be chosen from the closest limiting label
of $T$ for each subset. A non-photorealistic image is shown in Figure \ref{others}.

\subsection{Comparison}
We compare our model with contemporary methods
that deal with defocus modeling and post-capture scene refocusing. Previous
methods either use specific information such as precise knowledge of focus positions and/or scene
depth and mostly do not deal with fine characteristics of 
wide-aperture images. We provide a quantitative comparison wherever possible. We compare our method with 
that of \cite{Seitz, Levoy} for quality of refocusing, \cite{barron} for dual-pixel capabilites and \cite{Hach}
for quality of defocus kernels.

\paragraph{Comparison of Refocused Rendering}
Jacobs \etal \cite{Levoy} synthesize novel depth-of-field images using focal stacks by
defining a target sensor defocus map and generating a rendering closest to that map using ray 
space analysis.
Suwajanakorn \etal \cite{Seitz} also show reconstruction of defocused images from focal stacks 
however very little details about the process of generation of the defocused images is discussed. 
A direct comparison with these methods is non-trivial because it requires the ground-truth focus distance
for each shot using precisely engineered camera hardware that is not commonly available.
Our method compares favourably in the sense that it is independent of the capture-time focus distance
as we estimate the focus profile for the scene directly from the focal slices. In addition, we also make use of dual-pixels
to accurately reconstruct focus effects at depth-edges as shown in Figure \ref{dualPeaks}. Our method is therefore scalable
to generic focal stacks from any camera.

\paragraph{Comparison with Alpha-Blending}
Alpha-blending of different focal slices can be applied to generate refocused
renderings, as described in Barron \etal \cite{barron}. Such methods usually assume an 
extended canvas of background pixels which are blurred and blended to create a refocused
composite. Our dual-focus pixel method identifies the pixel intensity better and thus the textural 
information at partially occluded background pixel is encoded in our representation.
Such awareness of true pixel intensity and color is lost in the mutually exclusive focus textures that
are used in blending based methods. In Figure \ref{blendComp}, we demonstrate the quantitative benefit 
of our method over the background-stretched method of \cite{barron}. Background textural
details becoming apparent through a blurred foreground is clearly visible using our approach.

\paragraph{Comparison of Bokeh Quality}
Hach \etal \cite{Hach} render bokeh highlights using precise modeling 
and reconstruction of the PSF for a high-quality camera. They tediously calibrate 
an RGBD camera and learn the distribution of PSFs in order to simulate
DoF effects in a post-capture manner. They also describe a kernel stitching algorithm to
handle partial occlusions. While their model is designed for a high-end camera with a 
high dynamic range, we have shown a model which is applicable in a general setting, which works in 
the presence or absence of depth information. We compute the size and shape of defocus
kernels relative to the pixel location in the image and we also apply an intensity scaling 
to bokeh pixels (Sections V,VI). This leads to natural bokeh effects in
our refocused renderings. While the defocus kernels we estimate may not be
perfectly shaped for an arbitrary camera, the shortening of kernels due to vignetting and the
intensity scaling method creates photo-realistic bokeh, shown qualitatively 
in Figure \ref{showbokeh}.

\section{Acknowledgement}
This work is under consideration at the Elsevier Computer Vision and Image Understanding Journal.

\section{Conclusions}
In this paper, we propose a robust model to estimate the focus and defocus properties of a scene 
from a set of multi-focus images. We propose a composite measure of focus that reliably locates
in-focus pixels. We also present a geometric algorithm for refocusing from first principles, preserving
the fine effects of focus in challenging geometric situations. Our algorithm renders by distributing each 
pixel`s intensity to its neighbors using geometrical kernel sizes and shapes. Pixels close
to depth-edges and those with saturated intensities are also accounted for. We show the qualitative
and quantitative impact of our representation and refocusing algorithm. Our approach is ideally suited for 
image editing toolkits requiring precise manipulation of multi-focus imagery.

\bibliographystyle{splncs}
\bibliography{egbib}

\end{document}